\documentclass[10pt,journal,compsoc]{IEEEtran}
\ifCLASSOPTIONcompsoc
\else
  \usepackage{cite}
\fi
\ifCLASSINFOpdf
  \usepackage[pdftex]{graphicx}
\else
\fi
\usepackage{amsmath}
\usepackage{array}

\ifCLASSOPTIONcompsoc
  \usepackage[caption=false,font=footnotesize,labelfont=sf,textfont=sf]{subfig}
\else
  \usepackage[caption=false,font=footnotesize]{subfig}
\fi
\usepackage{dblfloatfix}
\usepackage{url}

\usepackage{hyperref}

\usepackage{booktabs}
\usepackage{xcolor}
\usepackage{multirow}
\usepackage{ascii}
\usepackage{makecell}
\usepackage{float}
\usepackage{xfakebold}
\usepackage{tikz}
\def\checkmark{\tikz\fill[scale=0.3](0,.35) -- (.25,0) -- (1,.7) -- (.25,.15) -- cycle;} 

\usepackage{diagbox}
\def\BibTeX{{\rm B\kern-.05em{\sc i\kern-.025em b}\kern-.08em
    T\kern-.1667em\lower.7ex\hbox{E}\kern-.125emX}}
    
\newcommand{\fbseries}{\unskip\setBold\aftergroup\unsetBold\aftergroup\ignorespaces}
\makeatletter
\newcommand{\setBoldness}[1]{\def\fake@bold{#1}}
\makeatother

\usepackage[style=ieee]{biblatex}
\begin{document}
%
\title{Data Leakage and Evaluation Issues in Micro-Expression Analysis}
%
%
%
%

\author{Tuomas~Varanka,~
        Yante~Li,~
        Wei~Peng,~
        and~Guoying~Zhao$^*$,~\IEEEmembership{Fellow,~IEEE}
\IEEEcompsocitemizethanks{\IEEEcompsocthanksitem $^*$ Corresponding author
\IEEEcompsocthanksitem T. Varanka, Y. Li and G. Zhao are with the Center for Machine Vision and
Signal Analysis, University of Oulu, Oulu, FI-90014, Finland.\protect\\
E-mail: firstname.lastname@oulu.fi
\IEEEcompsocthanksitem W. Peng is with the Department of Psychiatry \& Behavioral Sciences, Stanford University, CA 94305, United States. \protect\\
E-mail: wepeng@stanford.edu
}

\thanks{This work has been submitted to the IEEE for possible publication. Copyright may be transferred without notice, after which this version may no longer be accessible.}}
%
%

\markboth{Journal of \LaTeX\ Class Files,~Vol.~14, No.~8, August~2015}%
{Shell \MakeLowercase{\textit{et al.}}: Bare Demo of IEEEtran.cls for Computer Society Journals}
%



\IEEEtitleabstractindextext{%
\begin{abstract}
Micro-expressions have drawn increasing interest lately due to various potential applications. The task is, however, difficult as it incorporates many challenges from the fields of computer vision, machine learning and emotional sciences. Due to the spontaneous and subtle characteristics of micro-expressions, the available training and testing data are limited, which make evaluation complex. We show that data leakage and fragmented evaluation protocols are issues among the micro-expression literature. We find that fixing data leaks can drastically reduce model performance, in some cases even making the models perform similarly to a random classifier. To this end, we go through common pitfalls, propose a new standardized evaluation protocol using facial action units with over 2000 micro-expression samples, and provide an open source library that implements the evaluation protocols in a standardized manner. Code is publicly available in \url{https://github.com/tvaranka/meb}.
\end{abstract}

}

\maketitle

\IEEEdisplaynontitleabstractindextext

%
\IEEEpeerreviewmaketitle

\IEEEraisesectionheading{\section{Introduction}\label{sec:introduction}}

%
%
%
%
\IEEEPARstart{M}{icro-expressions} (MEs) are rapid and involuntary facial muscle movements, which may occur under high-stakes \cite{ekman:1969}. Compared to typical every day facial expressions (macro-expressions), MEs differ in three different ways. Micro-expressions only last a maximum of 0.5 seconds (there is some variation \cite{matsumoto2011evidence, yan2013fast}), while macro-expressions last longer. The second differentiating factor is the subtlety of the expression. Recently, \cite{micro_femg} showed that facial muscle contraction is significantly lower for ME. Lastly, the source of facial movement is involuntary for MEs, whereas macro-expressions are often used explicitly or \textit{voluntarily} to express an affective message.

It is the involuntariness that makes MEs so interesting. Several applications that utilize ME analysis have been proposed \cite{ekman:1969, yee:2018, xiaobai:2015, me_insights_survey}. One of the first applications lies in psychotherapy, discovered already in 1969 by Ekman and Friesen \cite{ekman:1969}. The two scientists analyzed a video of a mental health patient trying to find traits of self-harm intentions. However, despite the patient seeming happy, on a closer inspection of analyzing the footage frame-by-frame, Ekman and Friesen found a subtle sign of anguish on the patient's face. When confronted later, the patient admitted having self-harming thoughts and was not indeed happy as previously observed. Other applications include education \cite{me_for_education}, depression detection \cite{me_for_depression_detection}, \textit{etc}. In education the subtle facial movements could be used to indicate, \textit{e.g.}, whether a student is confused or frustrated. Being able to better understand mentally ill patients when they have trouble voicing their thoughts is an example of a potential application. A deep understanding of emotions \cite{barrett2019emotional} and especially MEs is still lacking in the field of psychology. The duration \cite{matsumoto2011evidence, yan2013fast}, magnitude \cite{micro_femg}, the rate of occurrence \cite{burgoon2018microexpressions}, the neural mechanisms behind MEs \cite{matsumoto2011evidence} and connections of micro facial movements to emotions \cite{4dmicro} are still not fully understood. By being able to automatically analyze short-term facial expressions, longer-term analysis and deeper understanding could be achieved.

Recognizing MEs is however extremely difficult for humans in real-time and hence automatic systems have been developed. Hand-crafted methods using LBP-TOP \cite{original_lbp_top} and optical flow \cite{xiaobai:2015, mdmo} were the first to be applied. Due to the limited data (only around 250 samples per dataset), deep learning methods were not as popular until recently. Off-ApexNet \cite{offapexnet}, STSTNet \cite{ststnet} and SSSNet \cite{noisy_mer} use shallow networks with optical flow to combat the limited data. As well, serious efforts have been made to address the limited data from transfer learning \cite{yante_icmi, micro_macro_pretrain, mer_pretraining} and synthetic data generation \cite{mie_x, au_cgan, megc2021}. However, recently, we have spotted a worrying trend with extremely high yet unreliable performances reaching close to perfect performance and potential issues during evaluation when analyzing available source code.

Kapoor and Narayanan \cite{data_leakage} conducted a meta review of data leakage and reproducibility in machine learning based sciences. They found more than 17 fields with 329 articles affected by data leakage or similar issues. Data leakage refers to using information from the testing data during the training procedure, giving an overly optimistic evaluation result. We find similar problems in the field of micro-expression recognition: several articles \cite{neural_micro_expression,aman,fsb,tanet,ma_net,ice_gan,mmnet,capsule_net} that are potentially affected by data leakage in micro-expression recognition from 2019-2022. These articles may have a data leakage problem that gives a large positive bias to the results. However, as many papers do not provide open source code, it is difficult to verify whether they have an issue or not. The most common issue is using test data to determine a hyperparameter, that is, the number of epochs. We find that methods achieving close to 80 F1-Score, but in fact only reach a performance of around 50 F1-Score when the data leak issue is fixed. Another possible issue we find is feature extraction or preprocessing using test data. Both cases constitute an issue with data leakage, where data from the testing set is used in the training procedure, resulting in a large positive bias. The concern with data leakage is that it creates a misleading understanding of the capabilities of models.

Fragmented protocols are also a major concern, as noted in a recent surveys \cite{me_insights_survey, micro_expression_deep_survey, }. The frequently used F1-Score can be computed in several ways \cite{f1_loso, macrof1_macrof1}, which can create confusion as different researchers may use different metrics even without knowing it. The use of different datasets with varying evaluation strategies and different number of emotions, subjects and samples creates more confusion and difficulties. Combining the issues of data leakage with fragmented protocols, fair comparison of methods becomes almost impossible.

Scientific progress relies heavily on improving and building on top of previous work. With the above mentioned issues, this becomes incredibly difficult. Hence it is crucial to fix these problems. We analyze the data leakage and various protocols and point out generally acceptable implementations. To act towards more united protocols, we propose a new protocol, \textbf{CD6ME}, that consists of six ME datasets with over 2000 samples. Firstly, with the introduction of new ME datasets \cite{mmew, 4dmicro, casme3}, there is more ME data available than ever before. Secondly, most ME datasets are labeled with both emotional labels and facial muscle movements, termed AUs (action units) \cite{facs}. By combining the datasets and using AUs, problems with the inconsistency of the labels can be largely alleviated, as the datasets are annotated by standardized FACS (facial action coding system) \cite{facs} certified coders. Emotional labels such as \textit{happiness}, \textit{sadness}, \textit{surprise}, \textit{anger}, \textit{fear} and \textit{disgust} have been shown to be inconsistent even within a single dataset \cite{noisy_mer, objective_classes}, which is further amplified when using multiple datasets together. Hence, using AUs as the ground truth, no samples need to be removed due to label inconsistencies as in the case of combined dataset of MEGC2019 (micro-expression grand challenge) \cite{megc2019} protocol and more datasets can be used compared to CDMER (cross-dataset micro-expression recognition) \cite{cdmer}.  Preliminary work on ME AU classification \cite{yante_icml, sca_yante, led} also supports the use of AUs over emotions.

To further steer into best practices, standardized protocols and to accelerate research, we offer an open source implementation library \textbf{MEB} for ME analysis. MEB implements tedious data loading routines, standardized training pipelines and multiple different models from the ME literature. All can be publicly accessed and tested. We hope that by providing an open source library we can avoid data leakage and fragmented evaluation protocols in the future and have more researchers share reproducible code from their experiments.

To summarize the contributions:
\begin{itemize}
    \item Common pitfalls found in the ME literature are showcased and discussed.
    \item A new composite cross-dataset action unit classification protocol for ME analysis is proposed.
    \item Comprehensive analysis is performed that compares action units and emotions in MEs.
    \item Open source implementation from all of the experiments and a library for ME analysis is provided for advancing open source, transparency and fair comparison with the proposed standardized protocol.
\end{itemize}

\section{Preliminary}
This section provides background on MEs and ME datasets that helps us understand the inconsistencies and limits in the performance evaluation of ME analysis, while also building a background for the proposed protocol. A comparison between AUs and emotional labels is made to advocate for the use of AUs for ME analysis.

ME analysis works as an umbrella term for any analysis done on MEs, be it recognition, spotting, classification or detection. The typical framework of a micro-expression analysis system consists of two phases: \textit{spotting} and \textit{recognition}. In the spotting phase, unsegmented videos are given as inputs and the task is to spot a temporal sequence during which an ME is occurring. Some ME spotting works \cite{yee:2018, spotting_benchmark} detect onset (the beginning of an facial expression), apex (maximum intensity of an facial expression) and offset (the ending of an facial expression). In the recognition phase, the pre-segmented video clip is classified to an emotion category such as \textit{happiness}, \textit{sadness}, \textit{surprise}, \textit{etc}. Recently, AU classification has been proposed to be used as an alternative to the recognition step. AU classification takes in a pre-segmented video clip and classifies which AUs are present in the clip.

\begin{figure}[!t]
    \centering
    \subfloat[All AUs]{\includegraphics[width=0.15\textwidth]{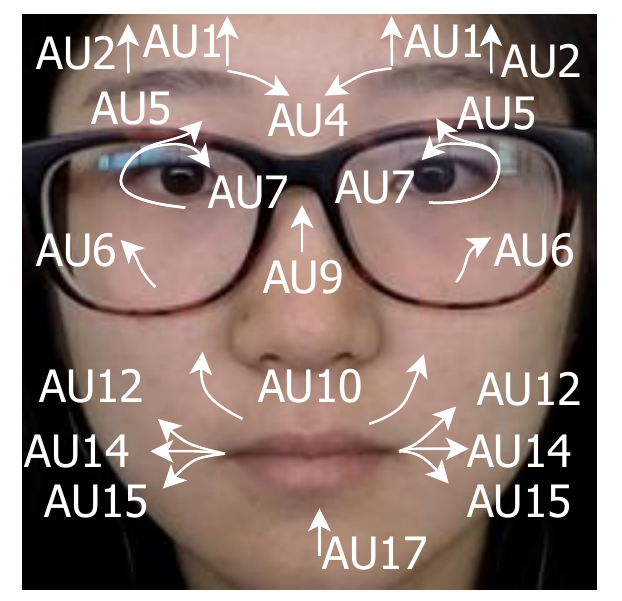}}
    \hfil
    \subfloat[Macro-Expression]{\includegraphics[width=0.15\textwidth]{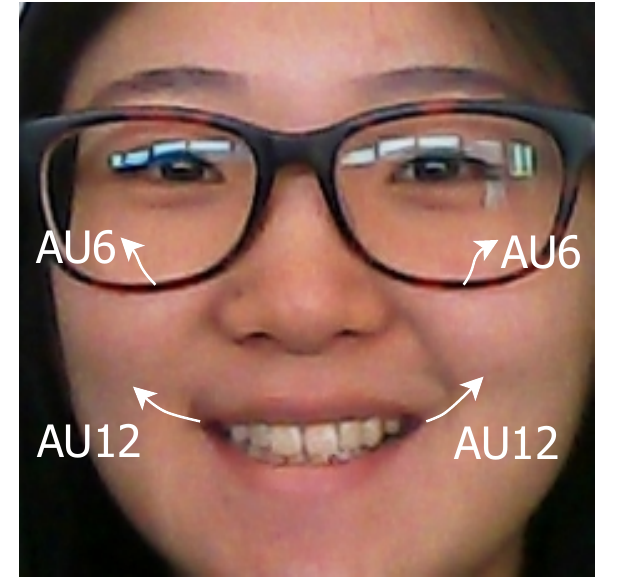}}
    \hfil
    \subfloat[Micro-Expression]{\includegraphics[width=0.15\textwidth]{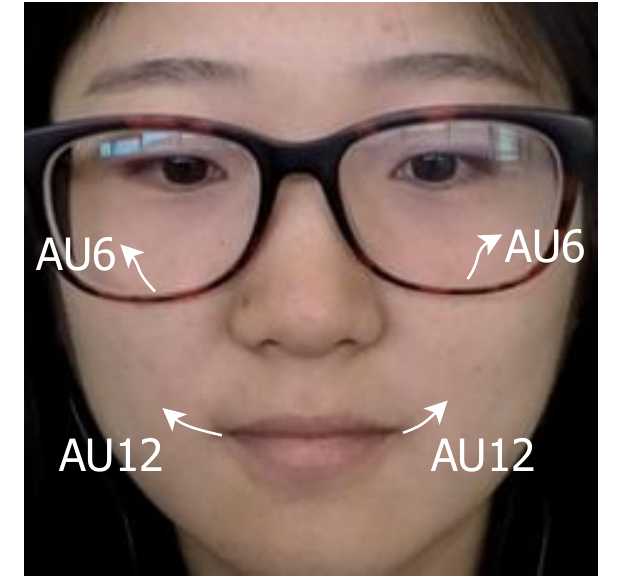}}
    \vfill
    \subfloat[Definitions of AUs]{\includegraphics[width=0.4\textwidth]{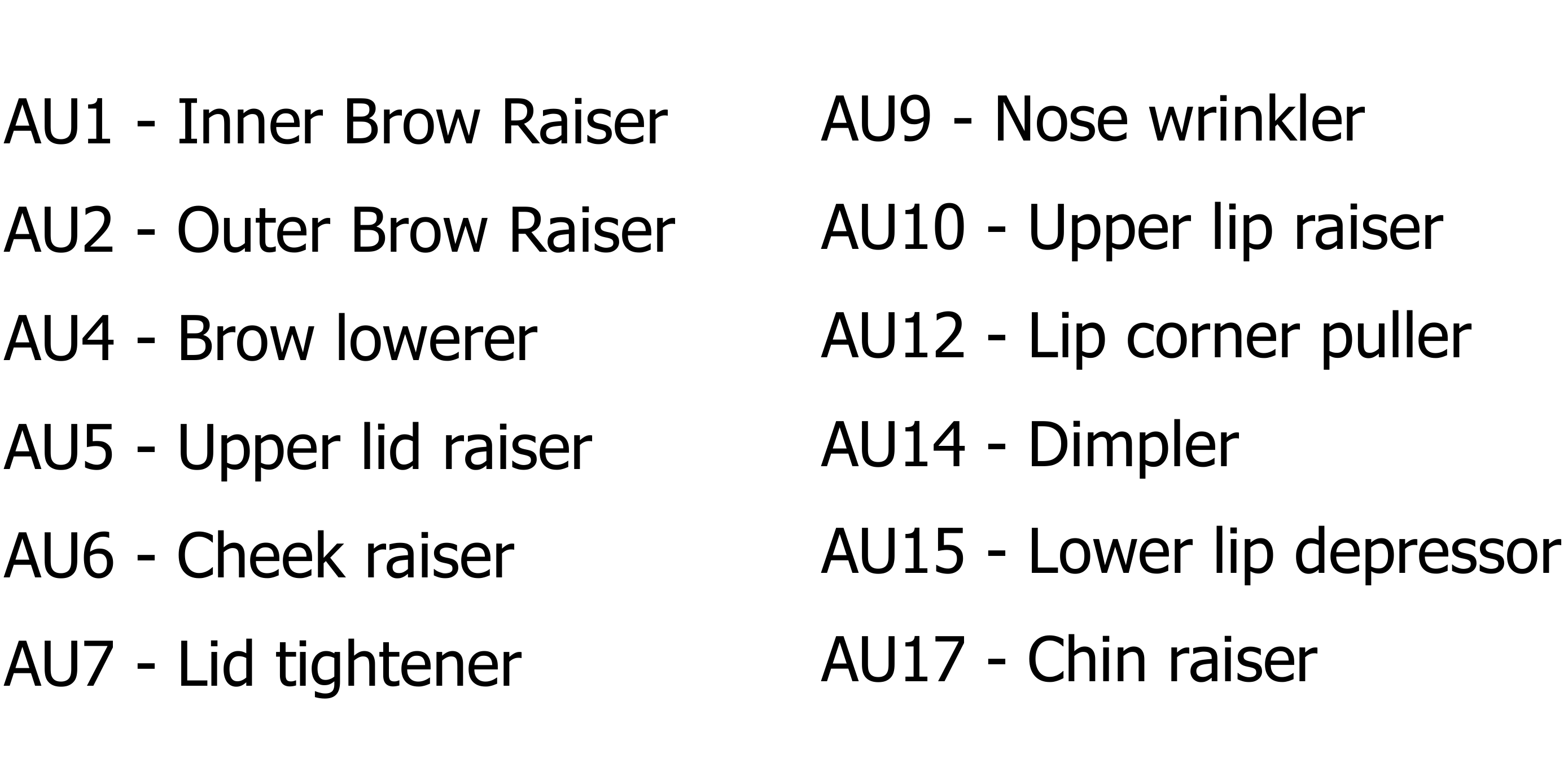}}
    \caption{Samples from MMEW dataset \cite{mmew} and definitions of AUs. Figures (b) and (c) show the difference between macro- and micro-expressions in magnitude of facial movements.}
\label{fig:samples}
\end{figure}

\begin{table*}[htp]
    \caption{Micro-expression datasets}
    \centering
    \begin{tabular}{|c|c|c|c|c|c|c|c|c|c|c|}
        \hline
        Dataset & Samples & Subjects & FPS & Facial resolution & $^1$ Emtn cls  & AUs & $^2$ AUs $\geq$ 10  & \#AUs & AU cardinality & Reliability \\
        \hline
        \hline
        CASME \cite{casme} & 189 & 19 & 60 & $150\times 190$ & 8 & 21 & 8 & 254 & 1.34 & 0.830 \\
        \hline
        CASME II \cite{casme2} & 256 & 26 & 200 & $280\times 340$ & 7 & 19 & 11 & 373 & 1.46 & 0.846 \\
        \hline
        SAMM \cite{samm} & 159 & 29 & 200 & $400\times 400$ & 8 & 27 & 6 & 226 & 1.42 & 0.820 \\
        \hline
        MMEW \cite{mmew} & 300 & 30 & 90 & $400\times 400$ & 7 & 21 & 12 & 500 & 1.67 & \textit{N/A} \\
        \hline
        4DME \cite{4dmicro} & 267 & 42 & 60 & $260\times 460$ & 5 & 19 & 8 & 454 & 1.70 & 0.75 \\
        \hline
        CASME3 \cite{casme3} & 860 & 94 & 30 & $280\times 340$ & 7 & 24 & 15 & 1077 & 1.25 & $^3$ 0.71  \\
        \hline
    \end{tabular}
    \label{tab:datasets}
    \\
    \vspace{0.2cm}
    $^1$ Number of emotion classes
    $^2$ Number of AUs that have at least 10 instances.
    $^{3}$ Measured from 7 classes of objective emotions.
\end{table*}

\subsection{Action units}
The FACS (facial action coding system) \cite{facs} is a taxonomy of fine-grained facial configurations. AUs (action units) constitute as a basic unit in encoding facial muscle movements. Figure \ref{fig:samples} shows examples of AUs from the MMEW \cite{mmew} dataset for both macro- and micro-expressions. For example, the prototypical expression of happiness can be decoded as a combination of AU6 (cheek raiser) and AU12 (lip corner puller) (See (b) and (c) from Figure \ref{fig:samples}). AUs can be considered as sign judgement of the face \cite{au_survey}, as opposed to emotional labels that attempt to convey the meaning. Due to this difference, automatic AU systems can be applied to a wider set of applications such as pain detection and analysis of nonaffective facial expressions \cite{au_survey}. FACS contains thirty action units that are related to facial muscles anatomically and 14 are unspecified miscellaneous actions \cite{au_survey}. Each AU can be given five different intensity levels (and one for neutral) denoted by an uppercase letter from A to E, where A is a trace and E is maximum \cite{facs}.

\subsection{ME Datasets}
\label{sec:datasets}
Currently, to our knowledge, there are a total of nine publicly available spontaneous ME datasets \cite{smic, casme, casme2, casme3, samm, mmew, 4dmicro, casme_squared, meview}. For a more detailed review of the datasets, see surveys \cite{mmew, micro_expression_deep_survey}. We summarize the most important information for this article from six datasets used in the proposed protocol in Table \ref{tab:datasets}. A subset of CAS(ME)$^3$ is chosen based on experimentation (see Appendix A from additional material). The subset contains only the MEs of part A of the dataset. It is referred as CASME3 for simplicity in the rest of the article. Each dataset contains front-facing videos of subjects from a laboratory setting. Facial expressions were elicited by emotion-inducing videos. Most datasets use a different set of emotion inducing videos. For example, SAMM \cite{samm} uses tailored videos for each subject for optimal inducement potential, while the CASME series \cite{casme, casme2, casme3} databases use a set of 13 selected emotional video clips. A high-stakes situation is used to facilitate the occurrence of spontaneous MEs \cite{ekman2}. Some variation in creating the high-stakes environment exists in the form of either a penalty \cite{4dmicro} or a reward \cite{samm} across the datasets. Each sample is a sequence of frames typically around 0.2 - 0.5 seconds long, that has been labeled with an emotion label and AUs.

The column \textit{AUs} in Table \ref{tab:datasets} refers to the different numbers of AUs that are annotated in the dataset. We can see that most datasets have around twenty different annotated AUs. The next column refers to number of AUs that have at least ten samples. We can see a discrepancy between the number of total AUs and ones with at least ten samples. Especially SAMM loses many of its AUs when this criterion is used, largely due to the also limited number of samples. \textit{\#AUs} refers to the total number of AU sequences in the dataset and cardinality computes the average number of AUs each sample has. Although CASME3 has three times the samples compared to 4DME, the number of AU sequences is only double.

\subsection{Annotation}
\label{sec:labeling}
Understanding the annotation of ME datasets is important, as the procedure is not as standard and trivial as is the case with common image datasets such as CIFAR \cite{cifar} and ImageNet \cite{imagenet_cvpr09}. Not properly understanding the labeling of datasets can lead to misunderstandings about the extent of the data and capabilities of models. Li \textit{et al.} \cite{4dmicro} cover the annotation process and related issues in detail. For our purposes, we summarize the most important parts. 

Compared to the onset and apex frames, the offset frame is more ambiguous as faces do not necessarily fully return to a relaxed state. Multiple AUs may have overlapping time spans. The time spans may therefore be based on a single AU, even though multiple ones exist. There still exist ongoing debates about categorical emotional labels for macro-expressions, although some agreement exists. This may not be the case for MEs due to the low intensity, suppression of true feelings and consecutive momentary rapid changes.

The \textit{emotion} label (\textit{e.g.}, \textit{happy}, \textit{sad}, \textit{surprise}, \textit{anger}, \textit{disgust}, \textit{etc.}) found in the ME datasets is obtained by different means in different datasets. SMIC uses only self-reports, while CASME II uses a combination of self-reported emotions, eliciting video emotion and action units. More recent datasets such as MMEW, 4DME and CASME3 only use AUs, which are then mapped to emotions. Different annotation strategies create a discrepancy between the datasets and makes comparison between the datasets inconsistent.

The reliability
\begin{equation}
    R = \frac{2 \cdot AU(C1, C2)}{N_{au}}
\end{equation}
between two annotators $C1$ and $C2$ computes the consistency of their independent annotations. Here $AU(C1, C2)$ gives the number of AUs for which coders $C1$ and $C2$ agreed on, and $N_{au}$ is the total number of AUs for the sample. The measure is between zero and one, where one means complete agreement. See the last column of Table \ref{tab:datasets} for the reliability measure for ME datasets.

\subsection{Emotions and Action Units}
Due to the above issues related to annotation, inconsistencies and noisy labels have been observed in the ME data \cite{noisy_mer, objective_classes}. Objective classes \cite{objective_classes} based on action units have been suggested to avoid this problem. More recently, directly using action units \cite{sca_yante} have also been suggested.

Mappings between AUs and emotions have been proposed. One such a table for macro-expressions is given in the FACS guide \cite{facs}. For micro-expressions, several tables have been proposed \cite{4dmicro, mmew, objective_classes}. However, a large meta study of facial expressions \cite{barrett2019emotional} suggests that there is no one-to-one mapping between facial movements and emotions. The study suggests that facial expressions may not be enough information, but context may be needed to capture emotions. Macro-expressions were considered in the study and hence the results may not be directly generalizable to MEs.

\textbf{Experiment}
We perform a simple study where we build a classifier and try to predict the emotion given the AUs. We do this using all of the data, \textit{i.e.}, we do not split a training and testing test. This is done to see how well AUs and emotions are correlated. We use a gradient boosting classifier, which is fed with binary features of different AUs (the AU label sequence). See the results from Table \ref{tab:predict_emotion_from_aus}. Even with the available AUs, the predictions are not 100\% correct which is the case of objective classes. CASME3 with self-reported emotion only obtains a 32.3 F1-Score from eight emotional classes and 100.0 F1-Score for the objective class. This supports the findings of the meta-study \cite{barrett2019emotional} and that there are no one-to-one mappings between AUs and self-reported emotions for MEs. Note that the predictions are not fully comparable, as the datasets have different sets of AUs and different sets of emotions. However, it should give some indications of the relationship between them.

\begin{table}[htp]
    \centering
    \caption{Predicting emotions from AUs}
    \begin{tabular}{c|c}
        Dataset & $F1_{macro}$ \\
        \hline
        CASME & 82.6 \\
        \hline
        CASME II & 87.2 \\
        \hline
        SAMM & 96.1 \\
        \hline
        SAMM (objective) & 99.2 \\
        \hline
        MMEW & 95.8 \\
        \hline
        4DME (self-report) & 34.9 \\
        \hline
        4DME (video emotion) & 49.8 \\
        \hline
        4DME (objective) & 100.0 \\
        \hline
        CASME3 & 32.3 \\
        \hline
        CASME3 (objective) & 100.0 \\
        \bottomrule
    \end{tabular}
    \label{tab:predict_emotion_from_aus}
\end{table}

For simplicity, with the above table we want to show a single number that can capture the inconsistencies well. However, the metric is quite one dimensional and hence it would be important to look at the clashes between different AUs. By further analyzing the data, we can see that most of the datasets have inconsistencies with the AUs and emotions. For example, in CASME II there are a total of twenty samples with AU14 (only AU14, no other AUs are present). Of these twenty samples, two of them are marked as repression, three as happiness and 15 as "others" category. Another example is AU4 in CASME3. Samples with AU4 are marked with seven available emotions ranging from happy to anger. These inconsistencies makes training on emotions difficult, especially with small datasets.

\textbf{Limits to predictions}
Based on our findings, we believe that AUs provide a more consistent and reliable label compared to emotional labels. However, AUs are not without drawbacks either. The reliability measure for each dataset (see Table \ref{tab:datasets}) is around 0.7-0.8. This means that we cannot expect models to perform with an accuracy of 100\% as the ground-truth labels contain noise. Hence, critical real world automated applications should be approached with care currently.

\section{Evaluation Issues in ME recognition}
\label{sec:data_leak}
Data leakage can cause a multitude of problems. Readers have to be more skeptical and careful going through articles. Leakage can create overly positive results that other researchers want to follow, further spreading the problem. It can discourage research, as methods with data leakage obtain state-of-the-art results that are difficult to surpass. This may be further amplified as non-replicable publications tend to be cited more often than replicable ones \cite{serra2021nonreplicable}. In this section, we will go over common pitfalls found in the ME literature. These include data leakage, imprecise use of the F1-Score and evaluation strategies. Although the use of different F1-Scores or evaluation strategies is not inherently wrong, without mentioning the details (or using a different one in code) can lead to misleading results.

\subsection{Data Leakage}

In data leakage, information from the testing data leaks to the training data that is used to train the model, leading to overly optimistic evaluation. First observed to our best knowledge in the MEGC 2019 (Micro-Expression Grand Challenge) \cite{megc2019} by NMER (neural micro-expression recognizer) \cite{neural_micro_expression}. As providing open source code was one of the requirements of the challenge, the code can be publicly accessed and tested. From there we can see that early stopping\footnote{Early stopping uses a validation set to determine the number of epochs needed to train a model before overfitting \cite{early_stopping}.} is performed, but with the testing data rather than the validation data. This leads to data leakage, as testing data is used to obtain the best model. When observing code, the data leakage can often be found in the training loop of an "if statement" where the current F1-Score is compared to the best F1-Score over different epochs.

We speculate that this practice of incorrectly using testing data for determining the number of epochs in K-fold validation likely comes from the field of image recognition where the datasets are extremely large and there is no need for cross-validation. Rather than running multiple experiments with different numbers of epochs, researchers often validate after every epoch. This gives the same result, but at a fraction of the cost. For cross-validation, this strategy is not applicable as cross-validation aims to estimate the generalization of the model and early stopping would result in multiple different epoch values. Each model with early stopping epoch value based on the testing data of a single fold would be optimal only for the specific fold, hence not necessarily generalizing to all folds. Using information from the test data during training can lead to a large positive bias, but the positive bias is misleading and not representative of the generalizable performance, especially when a fold is just a single subject. It is also helpful to think from the perspective of the final application why early stopping using testing data is not applicable to cross-validation. During the inference of real video clips, it is not possible to test which epoch model gives the best result.

We find 22 articles \cite{ capsule_net,aman,fsb,tanet,ma_net,ice_gan,mmnet,merastc,xia2021micro,Lei_2021_CVPR,fr_me,zhao2021two,li2021multi,zhao2021micro,zhang2021short,8918771,aouayeb2022spatiotemporal,zhao2021micro2,9325728,chen2021dffcn,aouayeb2021micro, rppg_me} that compare their results to a method with a data leak, namely NMER \cite{neural_micro_expression}. There are eight articles \cite{neural_micro_expression,aman,fsb,tanet,ma_net,ice_gan,mmnet, capsule_net} that publicly share their code and are likely to have an issue with data leakage.

\textbf{Experiments}
We reproduce (experimental details are described in Section \ref{sec:experiments} and in \cite{megc2019}) three different methods from the MEGC2019  competition in Table \ref{tab:nmer_data_leak}. Bold denotes the original implementation. We test the methods using two different training strategies: no early stopping (fixed epoch, same for all subjects) and with early stopping using test data (ES). By comparing the results of NMER with ES and the absence of ES, we can observe that ES provides a large positive bias to the result. The performance of NMER decreases from 77.2 to 49.6 in terms of $F1_{macro}$ when no ES is used. For Off-ApexNet and STSTNet we can see that their performances increase when using ES.

We further experiment whether using early stopping properly, \textit{i.e.}, by using validation data (ESV), has an impact. In each fold, we split the data into three categories. We first extract the testing data from the current subject's data. Then the remaining data are split by 80/20 to training and validation data. Early stopping is performed on the validation data, and testing data is not touched until the model has been fully trained. The results on NMER ESV show no improvement. The experiments show that using early stopping with test data can create a large positive bias, while using the validation data shows barely no impact.

\begin{table}[htp]
    \centering
    \caption{Early stopping comparison on the MEGC2019 task, $F1_{macro}$. Bold denotes original method in competition}
    \begin{tabular}{|c|c|c|c|c|c|c|}
        \hline
        Technique & ES & ESV & Full & SMIC & CASME II & SAMM \\
        \hline
        \textbf{NMER} \cite{neural_micro_expression} & \checkmark & & 77.2 & 72.8 & 83.4 & 71.7 \\
        \hline
        NMER \cite{neural_micro_expression} & & \checkmark & 49.6 & 48.4 & 50.4 & 33.8 \\
        \hline
        NMER \cite{neural_micro_expression} & & & 49.6 & 51.0 & 49.9 & 30.5 \\
        \hline
        \hline
        STSTNet \cite{ststnet} & \checkmark & & 77.4 & 73.3 & 87.3 & 69.7 \\
        \hline
        \textbf{STSTNet} \cite{ststnet} & & & 71.2 & 71.3 & 70.3 & 66.1 \\
        \hline
        Off-Apex \cite{offapexnet} & \checkmark & & 83.8 & 78.1 & 92.8 & 78.9 \\
        \hline
        \textbf{Off-Apex} \cite{offapexnet} & & & 70.3 & 68.3 & 77.4 & 60.5 \\
        \hline
        
    \end{tabular}
    \label{tab:nmer_data_leak}
\end{table}

\subsection{Other possible data leaks}
A common issue with data leakage, as noted in \cite{data_leakage}, is data leak during preprocessing, \textit{e.g.}, using testing data to determine normalization values or other hyperparameters. With the added complexity of using K-fold evaluation, such issues are more prone to occur without one's knowledge.

For instance, GA-ME \cite{ststnet_ga} uses a genetic algorithm (GA) for feature selection. Features are first extracted for all samples using a pre-trained model, after which the feature selection and classification are performed inside individual folds. This may however induce a data leak if the pre-trained model is trained using the same data as the evaluation is done on. To avoid the above issue, the pre-training should be done using additional data not part of the evaluation data or the pre-training should be done inside the individual folds.

RCN (recurrent convolutional network) \cite{rcn} is the first to use NAS (neural architecture search) for ME analysis. The authors optimized in an RCN search space to explore the best combination of different function modules. However, similar to above, a data leak may occur if the NAS is performed using the same data as is used in evaluation. Similar solutions from above can be applied.

Recently, generation of synthetic ME samples has been proposed as a potential solution to solve the low data problem \cite{mie_x, au_cgan, megc2021}. Generative models can be used to generate realistic looking ME samples. However, care should be taken when evaluating the effects of such synthetic data. If the evaluation is done with the same dataset that the generative model was trained on, a data leak may occur. When generated data is used for increasing training data size, caution should be paid to making sure that data leaks do not exist.

\subsection{Various F1-Scores}
\label{sec:f1_scores}
Due to the imbalanced data in ME datasets, accuracy provides a biased view of the performance of the model. A dummy model that always predicts the class with the most common occurrence could achieve good performance with accuracy. Use of F1-Score is a standard practice in the ME recognition task \cite{megc2019}. The F1-Score is defined as
\begin{equation}
    F1_{pr,re} = 2 \frac{Pr \cdot Re}{Pr + Re},
    \label{eq:f1_pr}
\end{equation}
which is the harmonic mean between precision (Pr) and recall (Re). By plugging in the definitions of precision and recall the equation can be simplified as
\begin{equation}
    F1_{tp,fp} = \frac{2TP}{2TP + FP + FN}.
    \label{eq:f1_tp}
\end{equation}
The two definitions \ref{eq:f1_pr} and \ref{eq:f1_tp} are equivalent. Here TP, FP, FN refer to true positives, false positives and false negatives, respectively.

The F1-Score can be generalized to a multi-class setting by a few different strategies. The commonly used and provided options from the Scikit-Learn \cite{scikit-learn} library are \textit{micro}, \textit{macro} and \textit{weighted}. For imbalanced labels, such as the case with MEs,
\begin{equation}
    F1_{macro} = \frac{1}{C} \sum_{c=1}^C \frac{2TP_c}{2TP_c + FP_c + FN_c}
    \label{eq:f1_macro}
\end{equation}
provides a more robust metric, where $C$ is the number of classes. The $F1_{macro}$ is also known as the unweighted F1 (UF1). One should be aware that when computing the F1-Score as noted by Opitz and Burst \cite{macrof1_macrof1}, the averaging can be done in two ways, as shown in Equation \ref{eq:f1_macro} or by first aggregating over the classes to compute precision and recall and using Equation \ref{eq:f1_pr} to compute the F1-Score. The two methods can have different results, but the authors find that Equation \ref{eq:f1_macro} is more robust, especially in imbalanced datasets.

\subsubsection{F1 and K-fold}
What further complicates the computation of the F1-Score for micro-expression recognition is evaluation using K-fold cross-validation. Already noted in the MEGC2019 (Micro-Expression Grand Challenge) \cite{megc2019, f1_loso}, this can be an issue. A common pitfall is to compute the F1-Score in each fold separately and aggregate the results together. This is the same issue as described above on how to compute the macro F1, but here over the folds rather than over the classes. Forman and Scholz \cite{f1_loso} show through experimentation that computing over the folds leads to a negative bias in the results.

\begin{figure}[htp]
    \centering
    \includegraphics[width=0.48\textwidth]{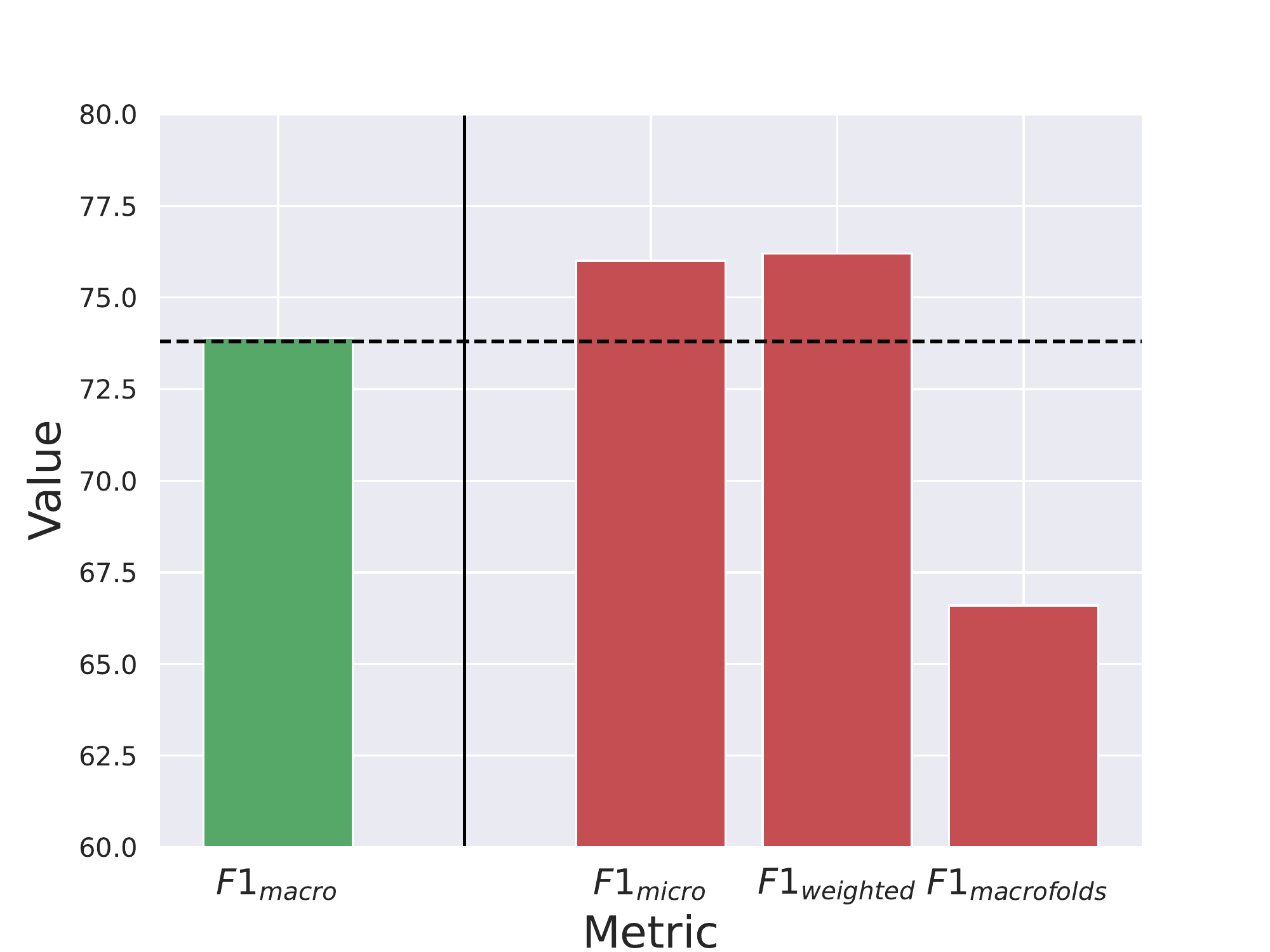}
    \caption{Results from computing F1-Scores differently on MEGC2019 with SSSNet. $F1_{macro folds}$ averages individually computed $K$ $F1_{macro}$ scores over the folds.}
    \label{fig:f1_scores}
\end{figure}

We find that issues with computing the F1-Score are less prevalent than the issue with data leakage. In Figure \ref{fig:f1_scores}, we show experimentation with different F1-Scores using SSSNet on the MEGC2019 dataset (see Section \ref{sec:experiments} and \cite{megc2019} for more details). As can be seen, both micro- and weighted F1 give a positive bias as they do not take the class imbalance into account. While averaging over the folds leads to a significant negative bias.

\subsection{Validation Strategy}
Verma et al. \cite{me_insights_survey} provide an excellent survey on different validation strategies used in ME recognition, which we summarize here. They split the validation strategies to three categories 1) person dependent evaluation (PDE), 2) person independent evaluation (PIE) and 3) cross domain evaluation (CDE). They note that most evaluations are using PIE. Through experimentation they show that PDE obtains a higher performance across all models over PIE. It is noted that PIE is a more reasonable evaluation strategy as it provides a more realistic setting for real world situations. Further, although limited studies on CDE have been performed, it shows that this is the hardest, but also the most suitable for real world settings.

In addition to different evaluation strategies, the number of samples and the number of used emotions may be different across articles. These small changes create confusion and obstruct direct comparison between models.

\section{CD6ME}
Evaluation across different micro-expression works has been fragmented. Different works typically use the commonly used datasets SMIC \cite{smic}, CASME II \cite{casme2}, SAMM \cite{samm} and MEGC2019 composite dataset \cite{megc2019}. However, different works use changing subsets with different number of emotions and samples. Add this to the common pitfalls discussed in the previous section and the comparison of different works is extremely difficult.

\begin{table*}
\caption{AU distribution of individual datasets and CD6ME}
\centering
\begin{tabular}{ll|rrrrrrrrrrrr|r}
    \toprule
    Dataset & Abbreviation & AU1 & AU2 & AU4 & AU5 & AU6 & AU7 & AU9 & AU10 & AU12 & AU14 & AU15 & AU17 & Total \\
    \midrule
    CASME & C1 & 23 & 17 & 70 & 0 & 1 & 4 & 40 & 3 & 9 & 23 & 14 & 13 & 217 \\
    CASME II & C2 & 26 & 22 & 130 & 2 & 13 & 39 & 13 & 16 & 34 & 27 & 16 & 25 & 363 \\
    CASME3 & C3 & 153 & 128 & 274 & 44 & 8 & 27 & 50 & 10 & 15 & 193 & 6 & 20 & 928 \\
    4DME & 4D & 46 & 54 & 102 & 8 & 25 & 89 & 1 & 2 & 53 & 4 & 6 & 11 & 401 \\
    MMEW & MM & 50 & 41 & 109 & 70 & 10 & 47 & 8 & 38 & 37 & 17 & 6 & 11 & 444 \\
    SAMM & SA & 6 & 18 & 23 & 10 & 3 & 46 & 5 & 6 & 28 & 13 & 4 & 6 & 168 \\
    \midrule
    CD6ME & - & 304 & 280 & 708 & 134 & 60 & 252 & 117 & 75 & 176 & 277 & 52 & 86 & 2521 \\
    \bottomrule
\end{tabular}
\label{tab:cross_dataset}
\end{table*}

\subsection{Construction of the Composite Dataset}
We propose a new cross-dataset protocol \textbf{CD6ME} to standardize the evaluation metrics and strategies. We consider the issue of small data by combining publicly available datasets together. The use of AUs allows us to combine the datasets as the annotation of AUs is standardized by having the annotators be qualified FACS coders. Furthermore, by using AUs, we can ameliorate the issue of ambiguous labels. By using a cross-dataset protocol, we ensure that the performance evaluation focuses on the capability of generalizing and dealing with domain shifts. Moreover, using a cross-dataset protocol reduces biases related to gender, ethnicity and age. As facial-expressions and emotions are highly influenced by culture \cite{barrett2019emotional}, only using samples from a single or few ethnicities like the CASME \cite{casme, casme2, casme3} datasets can lead to biases, which can also be avoided with cross-dataset protocol for datasets including subjects from different ethnicities.

We choose six publicly available ME datasets for our protocol: CASME \cite{casme}, CASME II \cite{casme2}, SAMM \cite{samm}, MMEW \cite{mmew}, 4DME \cite{4dmicro} and CASME3 \cite{casme3}. From the previously mentioned nine datasets in Section \ref{sec:datasets}, we discard SMIC due to its lack of AUs, CAS(ME)$^2$ and MEVIEW due to their small sample size. Twelve AUs are chosen on the basis of number of AU sequences: AUs 1, 2, 4, 5, 6, 7, 9, 10, 12, 14, 15 and 17 (See Figure \ref{fig:samples} for visualizations). Each of the chosen AUs should have at least 50 samples from all six datasets and at least one sample of an AU should be present in most datasets. The distribution of AUs for CD6ME can be seen from Table \ref{tab:cross_dataset}. Compared to a macro-expression action unit dataset, BP4D \cite{bp4d} uses AUs 1, 2, 4, 6, 7, 10, 12, 14, 15, 17 and 23. The distribution of AUs is similar with small differences to the ones we have chosen. 

\subsection{Comparison to Similar Works}
\textbf{MEGC2018} \cite{megc2018} proposed a cross-dataset protocol with two datasets: CASME II and SAMM. The training is done on CASME II and testing on SAMM and vice versa. Objective classes are used to handle the difference between emotional labels. The protocol is however limited to only two datasets. A composite dataset protocol is also proposed, where CASME II and SAMM are combined and the evaluation is performed using LOSO (leave-one-subject-out).

\textbf{MEGC2019} \cite{megc2019} extends the composite dataset in MEGC2018 by adding SMIC to it. However, this creates an issue as SMIC only has three emotional labels in it. Therefore, some classes of CASME II and SAMM are combined and some samples are left out. Another issue is the inconsistency between how the labels were annotated. As discussed in Section \ref{sec:labeling}, the datasets have different annotation strategies and hence there are likely to be inconsistencies between the datasets.

\textbf{CDMER} (cross-dataset micro-expression recognition) \cite{cdmer} proposes a protocol with SMIC and CASME II. The protocol uses the different subsets of SMIC, namely HS (high speed), VS (visual) and NIR (near infrared). Training is performed on one of the datasets/subsets and tested on another. The protocol suffers from several problems. First, only two datasets are used. Second, it experiences the same issue as MEGC2019, where the emotional labels are annotated differently.  Third, many of the tasks are questionable, \textit{e.g.}, task 1: train on HS and test on VS, but HS is just a higher frame rate version of VS. The domain shift is minimal, especially when considering that most methods sample frames, meaning that there is no big difference between HS and VS for most methods. Lastly, the protocol uses the domain adaptation paradigm rather than domain generalization. Domain adaptation allows methods to access unlabeled target domain data, while domain generalization does not. This makes the models less generalizable to real-world scenarios, as target domain data is not always available.

\subsection{Metrics and protocol}
The task is cross-dataset micro-expression action unit classification. The proposed protocol introduces familiar techniques from previous evaluation protocols, but to avoid misunderstandings, details are given in this section.
\subsubsection{Cross-dataset evaluation}
We propose to evaluate using leave-one-dataset-out (LODO) protocol, which is similar to the commonly used leave-one-subject-out (LOSO) protocol in ME recognition. The training and testing is repeated for $n_D$ times, where $n_D$ refers to the number of datasets. First, a single dataset is chosen as the testing data and the remaining $n_D - 1$ datasets are used as training data. Then, the process is repeated such that all $n_D$ datasets are used as testing data. 

The protocol mimics a realistic situation where the model is trained on one set of data, but inference is done on a different set of data where the distribution is often different. LODO also ensures that the same subject does not appear in both training and testing at the same time, similar to LOSO. Furthermore, training and evaluation are significantly faster since only $n_D$ models need to be trained, compared to $n_S$ models in LOSO, where $n_S$ refers to the number of subjects. Comparing CD6ME to MEGC2019 only six models are trained as opposed to the 68 in MEGC2019. In fact we can compute how many times samples are being used by multiplying the number of subjects with the number of total samples: $68 \cdot 442 = 30,056$ for MEGC2019 and with the number of datasets and samples $6 \cdot 2031 = 12,186$ for CD6ME. Despite using more than 4 times the data, the computational cost is more than halved for CD6ME.

\subsubsection{Action unit classification}
As discussed in Section \ref{sec:labeling}, emotion labels may be unreliable and using AUs allows us to combine multiple datasets together. In AU \textit{detection} an unsegmented video clip is given with frame level labels. The task is to predict a binary multi-label whether an AU exists for each frame separately. In AU \textit{classification} a pre-segmented video clip is given with a single binary multi-label \cite{au_survey2}. The task is to predict whether an AU exists for the whole clip. Since ME datasets do not have AU annotations frame-by-frame, the task is AU \textit{classification}.

\subsubsection{Metric}
There are significantly more negatives than positives in AUs which leads to an imbalance. A metric that takes this information into account is needed. Commonly the F1-Score from Equation \ref{eq:f1_pr} is used for evaluation of binary tasks, such as AU classification \cite{au_survey, au_survey2, au_transformer}. We refer to it as binary F1 and denote it by $F1_b$. The $F1_{macro}$ defined in Section \ref{sec:f1_scores} is not used here as the task is binary and $F1_{macro}$ is defined for multi-class outputs. It should be noted that $F1_b$ should be computed from all folds together and not averaged over them, as shown in Figure \ref{fig:f1_scores}.

\section{Experimentation}
\label{sec:experiments}
This section contains the introduction of the MEB library for ME analysis, experimental details, experiments with using additional data and experiments with the proposed CD6ME protocol.

\subsection{MEB}
Implementing evaluation strategies is complex and prone to errors, as we have seen in Section \ref{sec:data_leak}. MEB is an open source library for ME analysis. It includes implementations for data loading routines and conducting experiments using standardized protocols. Inspiration and design principles are taken from the OpenMMLab \cite{openmmlab} and Lightning \cite{lightning} libraries. The library is kept lightweight to ensure accessibility, while still allowing customization when needed. Extending new datasets, metrics, training and evaluation strategies is made easy by using abstract base classes. MEB supports both lazy and preloading of data for improved performance.

\subsection{Implementation Details}
All of the results shown in the experiments are produced by us and no results have been taken directly from other papers to ensure reproducibility and fair comparison. We use MEB for all of our experiments. For most models we follow the settings provided in the original implementation.  All experiments in this section are run five times with different seeds and the mean is taken to ensure reproducibility. Important high level details are shown here, for details of implementations see the code.

\textbf{Datasets}
Missing or out of bounds onset/apex/offset information is added for several datasets. Face area in all datasets are cropped using dlib \cite{dlib}. Optical flow methods use the cropped videos. Only the onset and apex are used (for CASME3 see the supplementary material) to compute the optical flow using \cite{secrets_of_optical_flow} following works from \cite{offapexnet, ststnet, noisy_mer}. Optical strain \cite{ststnet} is added for all optical flow -based methods except Off-ApexNet \cite{offapexnet}, following the original implementations. LBP-TOP and SCA use TIM (temporal interpolation model) \cite{tim} to standardize length of frames to 50 for all samples. ResNets use an input size of $112 \times 112$ and ResNet18(2+1)D uses eight frames sampled uniformly (See the supplementary material for CASME3).

\textbf{Models}
Since the predictions from the final fully connected layer are raw outputs, they need to be transformed to actual predictions. A threshold is used: $p = 1$ if the output is positive and $p = 0$ otherwise, where $p$ is the predicted label. ResNet models are pretrained \cite{pytorch} and use Adam \cite{adam} with a learning rate of $1e^{-4}$ and a weight decay of $1e^{-3}$.

\begin{table}[b]
    \centering
    \caption{Benefits of additional data. Evaluation on C2, training on different sets of datasets using $F1_b$. Methods use optical flow, while methods marked with \EOT use RGB as input. \underline{Underline} denotes best performance.}
    \begin{tabular}{|c|c|c|c|c|c|c|}
        \hline
        \diagbox{Model}{Data} & \#Prms & C1 & +SA & +MM & +4D & +C3 \\
        \hline
        Off-ApexNet \cite{offapexnet} & 0.8M & 31.5 & 37.5 &  \underline{42.5} & 45.1 & 43.4 \\
        \hline
        SSSNet \cite{noisy_mer} & 1.3M & \underline{34.3} & \underline{43.0} & 41.7 & \underline{46.7} & \underline{45.5} \\
        \hline
        \hline
        ResNet10 \cite{resnet} & 5.4M & 11.3 & 21.1 & 30.4 & 35.6 & 39.6 \\
        \hline
        ResNet18 \cite{resnet} & 11.7M & 10.4 & 22.1 & 32.0 & 39.2 & 43.1 \\
        \hline
        ResNet34 \cite{resnet} & 21.8M & 13.9 & 23.0 & 33.0 & 37.2 & 42.1 \\
        \hline
        ResNet50 \cite{resnet} & 25.5M & 0.6 & 2.1 & 8.2 & 24.5 & 29.4 \\
        \hline
        ResNet101 \cite{resnet} & 44.5M & 0.4 & 0.2 & 7.7 & 21.4 & 27.6 \\
        \hline
        RN18(2+1)D \cite{resnet3d_2p1} \EOT & 31.5M & 8.4 & 7.8 & 18.1 & 24.7 & 33.9 \\
        \hline
    \end{tabular}
    \label{tab:large_benefits}
\end{table}

\begin{table*}[t]
\caption{Results on the cross-dataset protocol. The results are for individual action units over all of the different dataset folds with $F_b$ metric. \EOT marks methods that use RGB as input, while others use optical flow.}
\centering
\scriptsize
{
\centering
\begin{tabular}{|c|c|c|c|c|c|c|c|c|c|c|c|c|c|c|}
    \hline
     Type & Method & AU1 & AU2 & AU4 & AU5 & AU6 & AU7 & AU9 & AU10 & AU12 & AU14 & AU15 & AU17 & Average\\ [0.5ex] 
    \hline
     \multirow{5}{*}{Hand-crafted} & Constant & 26.0 & 24.2 & 51.7 & 12.4 & 5.7 & 22.1 & 10.9 & 7.1 & 15.9 & 24.0 & 5.0 & 8.1 & 17.8 \\
        \cline{2-15}
         & LBP-TOP \cite{original_lbp_top} \EOT & 41.6 & 36.7 & 62.0 & 0.0 & 0.0 & 0.0 & 1.7 & 0.0 & 0.0 & 3.5 & 0.0 & 0.0 & 12.1 \\
        \cline{2-15}
         & MDMO \cite{mdmo} & 59.4 & 57.0 & 72.2 & 16.7 & 2.3 & 18.1 & 10.9 & 0.0 & 27.4 & 25.5 & 13.0 & 27.8 & 27.5 \\
        \cline{2-15}
         & SVM + OF & 71.0 & 65.9 & 84.2 & 13.6 & \underline{10.6} & 44.4 & 17.2 & 10.3 & 26.7 & 39.1 & 18.2 & 32.7 & 36.2 \\
        \cline{2-15}
         & GA-ME \cite{ststnet_ga} & 71.1 & 67.6 & 85.7 & 1.1 & 0.0 & 26.2 & 7.8 & 2.1 & 12.8 & 15.4 & 0.0 & 34.7 & 27.0 \\
        \cline{1-15}
         \multirow{8}{*}{Deep learning} & Off-ApexNet \cite{offapexnet} & 74.9 & 70.2 & 86.3 & 13.5 & 3.3 & 44.5 & \underline{36.6} & 18.3 & 32.0 & 37.7 & 19.0 & 38.2 & 39.5 \\
        \cline{2-15}
         & STSTNet \cite{ststnet} & 74.2 & 68.8 & 85.7 & 3.6 & 6.4 & 29.0 & 18.6 & 9.0 & 16.7 & 29.0 & 11.8 & 28.3 & 31.8 \\
        \cline{2-15}
         & RCN-A \cite{rcn} & 74.8 & 71.0 & 85.3 & 4.7 & 0.0 & 24.1 & 15.9 & 0.0 & 21.1 & 23.7 & 0.0 & 24.0 & 28.7 \\
        \cline{2-15}
         & NMER \cite{neural_micro_expression} & 19.1 & 19.2 & 43.3 & 9.3 & 6.3 & 9.1 & 12.5 & 4.8 & 18.1 & 22.6 & 3.3 & 6.4 & 14.5 \\
        \cline{2-15}
         & SSSNet \cite{noisy_mer} & 74.6 & \underline{72.1} & \underline{87.5} & 13.6 & 5.3 & \underline{48.6} & 20.3 & \underline{19.3} & 36.2 & \underline{40.8} & 22.7 & \underline{44.7} & \underline{40.5} \\
        \cline{2-15}
         & ResNet10 \cite{resnet} & 68.7 & 65.2 & 83.8 & 8.3 & 6.4 & 39.6 & 11.0 & 6.8 & 31.0 & 29.6 & 8.8 & 39.6 & 33.2 \\
        \cline{2-15}
         & ResNet18 \cite{resnet} & \underline{76.2} & \underline{72.1} & 87.1 & 12.6 & 5.5 & 38.2 & 8.6 & 18.4 & \underline{37.3} & 33.7 & 12.4 & \underline{44.7} & 37.2 \\
        \cline{2-15}
         & ResNet34 \cite{resnet} & 72.1 & 71.7 & 85.8 & 11.2 & 5.4 & 38.7 & 16.0 & 13.5 & 36.5 & 39.9 & 14.8 & 39.2 & 37.1 \\
        \cline{2-15}
         & SCA \cite{sca_yante} \EOT & 42.3 & 42.8 & 56.2 & 14.8 & 1.2 & 23.4 & 13.4 & 3.0 & 21.8 & 37.1 & 4.0 & 22.9 & 23.6 \\
        \cline{2-15}
        & LED \cite{led} \EOT & 52.7 & 45.7 & 63.7 & 7.9 & 0.7 & 19.3 & 13.6 & 8.5 & 26.5 & 36.7 & \underline{33.0} & 31.7 & 28.3 \\
        \cline{2-15}
         & RNet18(2+1)D \cite{resnet3d_2p1} \EOT & 54.2 & 49.4 & 72.7 & \underline{26.9} & 5.2 & 20.4 & 11.7 & 12.4 & 23.3 & 25.5 & 9.7 & 41.9 & 29.5 \\
    \hline
\end{tabular}\par
}
\label{tab:benchmark_aus}
\end{table*}

\textbf{Loss function} Since the labels are binary multi-label the binary cross entropy with logits

\begin{equation}
\begin{aligned}
    \mathcal{L}(x, y) = \frac{1}{N_{au} \cdot N}\sum_{au=0}^{N_{au}}\sum_{n=0}^{N - 1} & y_n^{au} \cdot \log(\sigma(x^{au})) \\
     & + (1 - y_n^{au}) \cdot \log(1 - \sigma(x^{au})),
\end{aligned}
\end{equation}
is used as the loss function for each AU and the results are aggregated together. $\sigma(x) = \frac{1}{1 + e^{-x}}$ is the sigmoid function.

\subsection{Benefits of Composite Data}
Limited data is believed to be one of the biggest problems for ME analysis \cite{micro_expression_deep_survey, mmew}. We test this hypothesis to see whether the performance can be increased by using more data. Furthermore, we experiment whether larger models benefit from the addition of data. For the experiment, we choose CASME II as the testing data, while the training data is composed of different combinations of datasets. We choose two models from the ME literature and ResNet \cite{resnet}, as it is easy to scale to different sizes of models. The results can be seen in Table \ref{tab:large_benefits}. The training is started only using C1 (see Table \ref{tab:cross_dataset} for abbreviations) in the third column. Next, in the fourth column, SA is added, so now the training is performed with C1+SA. In the last column, all the datasets except for C2 are used.

Across all the models we can see improvements when using more data. Large improvements can be seen from the large networks like ResNet50, where the $F1_{b}$ increased from 0.6 to 29.4. In general, greater improvements are observed from using more data from the larger models. When using only a single dataset (column C1), which has been standard in previous ME analysis, we can see that larger models are not able to compete with specially designed smaller models. SSSNet is able to obtain an $F1_b$ of 34.3 while the best ResNet only reaches a performance of 13.9 and the extremely large networks fall far behind. When all datasets are used together, ResNet18 is as able to match Off-ApexNet in performance, and the performance gap between ResNets and ME models is notably smaller.

\subsection{Benchmark Experiments on CD6ME}
Experiments are conducted with a large number of different methods and the results are reported in Table \ref{tab:benchmark_aus}, which shows the results by each AU. See the supplementary material for results by each dataset. We choose methods from the ME literature and a set of ResNet models due to their success in related tasks. The ME methods are chosen based on  simplicity of implementation, availability of code and reproducibility. As mentioned previously, data leakage and evaluation issues are largely affected, which made reproducing results difficult. We split the methods based on hand-crafted and deep learning solutions. For most methods the input is optical flow, while methods marked with \EOT use RGB as their input.

Previous ME work has ignored simple baselines that give a proper context to the task. The \textit{constant} method always predicts the positive label, \textit{i.e.}, only 1s for every AU and every sample ($F1_b$ does not consider true negatives). Another overlooked method not previously experimented with to our knowledge is \textit{SVM+OF}, where SVM is used as the classifier with input being optical flow flattened to a vector.

\textbf{Results over AUs} Table \ref{tab:benchmark_aus} shows the results for each AU, where the results are computed from all folds. On average we can see that AU4 obtains the highest results. AU4 also has the largest number of samples, a total of 708, while the next most common AU1 only has 304. The number of samples is not the only explanation for good performance. AU14 has 277 samples, but on average achieves an $F1_b$ of around 25, while AU17 is closer to 30 despite only having 86 samples. The poor performance of AU14 can be explained by the similarity of low intensity actions of AUs 12, 14 and 15 being highly similar as also observed in \cite{4dmicro, led}.

\textbf{Results of methods over AUs}
The baseline method \textit{constant} achieves an $F1_b$ of just 17.8. The commonly used benchmark for ME recognition, LBP-TOP, receives a worse result. Another common baseline, MDMO, gets an $F1_b$ of 27.5. In the intra-dataset case the two methods achieve similar results \cite{noisy_mer}, the discrepancy may be explained by the use of optical flow. The simple SVM + OF obtains a surprisingly good performance, even outperforming some deep learning models. After not using early stopping in NMER, poor performance, even below the \textit{constant} method can be seen. SSSNet is able to obtain the best result from six out of twelve AUs. Interestingly, SVM + OF attains the best result of AU6, which is the most difficult AU in terms of performance from different methods. We hypothesize that this may be due to SVM only using single labels, as opposed to neural network architectures that use a multi-label approach. The small ME models, namely Off-ApexNet and SSSNet, obtain the best results with larger ResNet models being a few percentages behind. This reiterates the results from Table \ref{tab:large_benefits} and shows the viability of larger models.

Results in \cite{xu2022asymmetric} show that methods using RGB input do not perform well under the cross-dataset task compared to optical flow. In our experimentation they however perform on par with some optical flow methods. This is most likely explained by the increase in training data that forces the RGB methods to learn domain invariant features, as shown in Table \ref{tab:large_benefits}.

\subsection{Discussion}
While a previous study has suggested the use of AUs \cite{sca_yante}, the work does not provide detailed explanation on the choice and benefits. Research on cross-dataset \cite{cdmer} and composite datasets \cite{megc2019} have been done previously, but separately and in limited quantities. By combining the above together in this paper, we are able to evaluate methods in a more realistic setting, while providing increased performance by using additional data.

The use of optical flow has been very popular due to its empirical performance. However, the method is limited to the accuracy of optical flow and there may be information loss that is important for MEs. Multiple AUs may occur at different times, using only the apex may therefore miss one or more AUs. Optical flow is also likely to not work well in-the-wild scenarios with head movement and lighting changes. With the proposal of larger data, we encourage future research to the use of RGB video inputs. The results in Table \ref{tab:benchmark_aus} show promising results for the use of RGB as input when using large composite data.

Data-centric approach \cite{data_centric} refers to focusing on improving the input data rather than the model in machine learning. As shown by our work, significant gains can be obtained without touching the models. We encourage other researchers to focus more on the data, rather than the models. Not only the quantity, but also the quality.

\textbf{Limitations} Although the cross-dataset is a more realistic setting, the data is still from a laboratory setting, which limits the applicability for in-the-wild scenarios. Another limitation is the need for data which requires capturing spontaneous subtle facial-expressions from human subjects and accurate labor intensive annotations. The use of AUs requires knowledge of FACS and emotional information and can be difficult for non-experts. The experiments connecting AUs and emotional labels presented in this paper use limited data and there may be several confounding variables, which should be further explored in future..

\section{Conclusion}
In this paper we point out common pitfalls such as data leakage and fragmented use of evaluation protocols in micro-expression recognition. We show how these issues can be avoided and how they affect the results. We propose a new benchmark, CD6ME, that uses a cross-dataset protocol for generalized evaluation. Action units are used instead of emotional classes for a more objective and consistent label. We investigate the effect of using composite datasets for training and show a notable increase in performance across different models. A micro-expression analysis library, MEB, with the implementation of data loading routines, training loops and several commonly used micro-expression models, is introduced and openly shared.

\nocite{pytorch, pandas, numpy, matplotlib, scikit-learn, scipy}


%

\appendices

\section{CAS(ME)$^3$ Experiments}
\label{app:casme3}
In this section we provide experiments for our choice on only choosing part A of CAS(ME)$^3$ \cite{casme3} and further discussion on the dataset. 

CAS(ME)$^3$ contains three parts: A, B and C. A is the standard in laboratory setting, similar to other datasets. B contains unlabelled data. C contains a more realistic setting, where the emotion is induced by an interrogation rather than by watching videos. Part B is not included in our experiments as it does not contain label information. In Table \ref{tab:effects_of_casme} we experiment with including C3 (3rd row), C3AC (both part A and part C, 2nd row) and excluding C3 completely (1st row). The experiment is conducted with the same settings as in Table \ref{tab:large_benefits}, \textit{i.e.}, evaluation is performed on CASME II using SSSNet and datasets used during training are changed. Here $\emptyset$ refers to an empty dataset and $+$ to adding a dataset. It can be seen that both datasets hurt the performance with different magnitudes depending on the dataset configuration. As part A contains 860 samples and part C only 165 we decide to only keep part A as it contains a large number of samples and the performance drop is not as severe.

\begin{table}[htp]
    \centering
    \caption{How CASME3 affects training performance}
    \begin{tabular}{c|c|c|c|c|c}
        Datasets & + $\emptyset$ & + C1 & + SA & + MM & + 4D \\
        \midrule
        $\emptyset$ & N/A & 34.3 & 43.0 & 41.7 & 46.7 \\
        C3AC & 30.9 & 34.0 & 38.3 & 38.8 & 43.2 \\
        C3 & 33.1 & 34.2 & 40.0 & 41.3 & 45.5 \\
        \midrule
        \bottomrule
    \end{tabular}
    \label{tab:effects_of_casme}
\end{table}

We would like to further discuss details related to CAS(ME)$^3$ that impact the training and evaluation. Part A of CAS(ME)$^3$ uses a different criteria for annotation than other datasets. Other ME datasets use the criteria of $offset - onset \leq 500ms$ for MEs. CAS(ME)$^3$ add the following criteria $apex - onset \leq 250ms$ which is checked if the first criteria is not valid. The latter criteria is proposed in \cite{yan2013fast}. Some samples in CASME3 are long, one being over 24 seconds long. The sample is however considered an ME as the second criteria is met. When using video inputs, researchers should be careful when sampling frames as most of the important information is likely to be concentrated around the apex. In our experiments we first sample only the beginning frames from $onset$ to $2 \cdot apex - onset$. 

Another peculiarity to consider is that in some cases the onset and the apex are marked as the same frame. This is explained by the authors in \url{https://github.com/jingtingEmmaLi/CAS-ME-3} and is caused by frame drops and late recording. Optical flow methods use the optical flow between the onset and the apex, however in these cases it is not possible. For these cases we use the optical flow between the apex and the offset.

Some samples have fewer frames than eight, which is used by our ResNet18(2+1)D. These samples are interpolated to eight frames with trilinear interpolation.

\section{Objective label versus emotional label}
\label{app:objective_vs_emotional}
Additional experiments are conducted in continuation of Table \ref{tab:predict_emotion_from_aus} to understand whether the use of objective labels and emotional labels has an impact on performance. We also use additional data to see its effect on the performance. A similar study was also conducted in \cite{objective_classes}, which is extended here by using additional data during training and including more experiments from different datasets.

The experiments are conducted using the LOSO protocol with macro F1. Seven objective classes and depending on the dataset seven or eight emotional labels (CASME and SAMM use eight, while the rest use seven) are used. 4DME is not used due to its use of compound expressions. By additional data we mean using data from all the other datasets. For objective labels this is all of the data as we transform the AUs to objective labels according to \cite{objective_classes}. For emotional labels, samples which do not correspond to the testing dataset's emotions are discarded. This leaves total sample size to be around 1200-1500 depending on the testing dataset. All of the results are obtained by using SSSNet with optical flow with similar settings as in Section \ref{sec:experiments}.

Table \ref{tab:objective_vs_emotion} shows the results. The top part of the table shows results without using additional data (\textit{i.e.}, only samples within the dataset). In general, it can be observed that the objective label obtains a higher performance compared to the emotional label. For CASME3 it can be observed that the difference between objective and emotional label is largest. This result is supported by Table \ref{tab:predict_emotion_from_aus}, where the mapping from AUs to the emotional labels only achieves a 32.3 F1-score.

The bottom part of Table \ref{tab:objective_vs_emotion} shows results using additional data. For CASME3 no significant difference can be observed, most likely due to the dataset size already being large without the extra data. For CASME, CASME II and MMEW the results are similar. With objective label the performance is increased, while with emotional labels the performance decreases. This is likely explained by the consistency of objective labels, where all datasets use the same mapping from the AUs. Having consistent labels allows learning from other datasets. Emotional labels are annotated differently in each of the datasets, hence the additional data can in fact decrease the performance due to "noisy" labels.

For SAMM the performance increases in both cases by a large amount. We hypothesize that this may be due to the original dataset being more diverse and having a smaller sample size of only 159 samples. However, further investigation is required.

Comparing the results from this section to Table \ref{tab:large_benefits}, the performance increases are relatively mild when using additional data. This is likely due to the use of different protocol. With LOSO the methods have access to samples within the dataset, which may lead to overfitting and poor generalization.

\begin{table*}[htp]
    \centering
    \caption{Performance comparison between objective and emotional labels using LOSO validation with macro F1-score}
    \begin{tabular}{|c|c|c|c|c|c|c|}
        \hline
        Method & Extra data & C1 & C2 & SA & MM & C3 \\
        \hline
        Objective label & & 40.7 & 50.6 & 40.1 & 45.0 & 52.6 \\
        \hline
        Emotional label & & 37.0 & 44.6 & 29.0 & 44.8 & 24.0 \\
        \hline
        \hline
        Objective label & \checkmark & 42.4 (+1.7) & 53.3 (+2.7) & 57.3 (+17.2) & 49.8 (+4.8) & 51.8 (-0.8) \\
        \hline
        Emotional label & \checkmark & 31.4 (-5.6) & 43.7 (-0.9) & 35.5 (+6.5) & 40.8 (-3.9) & 24.4 (+0.4) \\
        \hline
    \end{tabular}
    \label{tab:objective_vs_emotion}
    \\
    \vspace{0.1cm}
\end{table*}

\section{CD6ME results across datasets}
Table \ref{tab:benchmark_dataset} shows the results from each fold of LODO (leave-one-dataset-out), where the individual F1-Scores of AUs are averaged together. These experiments correspond to Table \ref{tab:benchmark_aus}, but the results are averaged over the AUs. This showcases differences between datasets. It can be seen that the performance across different datasets is mostly similar with some variation. C2 achieves the best performance from most methods, with the best method obtaining 46.3. The lowest performance from the datasets, \textit{i.e.}, the most difficult dataset is 4D with 34.3. We can see that RGB methods often get the best results on C1. In fact, the best results on C1 is obtained by LED.

\begin{table*}[h!t]
\caption{Cross-dataset evaluation. The dataset refers to the dataset being tested, while the rest of the datasets are used for training. For example, in the first column, C1 is used for testing while C2, SA, 4D, MM and C3 were used for training. \EOT marks methods that use RGB as input, others use optical flow.}
\centering
\begin{tabular}{|c|c|c|c|c|c|c|c|c|c|c|c|} 
    \hline
    Metric & Type & Method & C1 & C2 & SA & 4D & MM & C3 & Average \\
    \hline
        \multirow{13}{*}{$F1_{b}$} & \multirow{5}{*}{Hand-crafted} & Constant & 16.1 & 19.4 & 15.3 & 20.2 & 20.7 & 15.1 & 17.8 \\
        \cline{3-10}
         & & LBP-TOP \cite{original_lbp_top} \EOT & 17.9 & 16.5 & 9.7 & 13.8 & 20.1 & 4.4 & 13.7 \\
        \cline{3-10}
         & & MDMO \cite{mdmo} & 29.5 & 31.7 & 21.1 & 27.5 & 29.1 & 23.3 & 27.0 \\
        \cline{3-10}
         & & OF + SVM & 31.9 & 42.8 & 38.7 & 31.5 & 31.6 & 33.1 & 34.9 \\
        \cline{3-10}
         & & GA-ME \cite{ststnet_ga} & 23.8 & 31.0 & 32.0 & 23.3 & 29.6 & 25.6 & 27.6 \\
        \cline{2-10}
         & \multirow{8}{*}{Deep learning} & Off-ApexNet \cite{offapexnet} & 30.5 & 43.1 & \underline{43.6} & 30.6 & 37.1 & \underline{36.8} & 37.0 \\
        \cline{3-10}
         & & STSTNet \cite{ststnet} & 29.9 & 38.3 & 34.9 & 23.1 & 30.2 & 29.2 & 31.0 \\
        \cline{3-10}
         & & RCN-A \cite{rcn} & 24.2 & 34.6 & 33.5 & 24.7 & 26.4 & 27.1 & 28.4 \\
        \cline{3-10}
         & & NMER \cite{neural_micro_expression} & 11.3 & 11.5 & 9.2 & 10.2 & 12.3 & 11.9 & 11.1 \\
        \cline{3-10}
         & & SSSNet \cite{noisy_mer} & 33.1 & \underline{46.3} & 42.3 & \underline{34.3} & \underline{38.1} & 36.2 & \underline{38.4} \\
        \cline{3-10}
         &  & ResNet10 \cite{resnet} & 21.2 & 39.8 & 36.7 & 33.8 & 32.6 & 27.7 & 32.0 \\
        \cline{3-10}
         &  & ResNet18 \cite{resnet} & 25.0 & 41.9 & 42.4 & 33.1 & 35.4 & 33.7 & 35.2 \\
        \cline{3-10}
         & & ResNet34 \cite{resnet} & 28.2 & 41.2 & 36.3 & 34.0 & 35.0 & 33.2 & 34.7 \\
        \cline{3-10}
         & & SCA \cite{sca_yante} \EOT & 27.0 & 29.3 & 19.2 & 17.2 & 25.7 & 16.0 & 22.4 \\
        \cline{3-10}
         &  & LED \cite{led} \EOT & \underline{37.5} & 38.6 & 22.3 & 33.9 & 26.8 & 16.7 & 29.3 \\
        \cline{3-10}
         & & RNet18(2+1)D \cite{resnet3d_2p1} \EOT & 27.6 & 30.0 & 33.0 & 21.8 & 34.4 & 21.4 & 28.0 \\
    \hline
\end{tabular}
\label{tab:benchmark_dataset}
\end{table*}

\section{Experiments with MEGC2018}
We provide additional experiments using the two protocols proposed in the MEGC2018 competition \cite{megc2018} in order to compare our proposed techniques to previous research. Two tasks A) and B) are used. Task A is the holdout-database evaluation, \textit{i.e.}, cross-dataset evaluation using two datasets, CASME II \cite{casme2} and SAMM \cite{samm} with five objective labels and uses AUR (unweighted average recall) as the metric. Task B is the composite database evaluation. It uses the the same datasets and labels as task A, but evaluation is done using LOSO (leave-one-subject-out) protocol and uses macro F1-score as its metric. The five objective labels are obtained by mapping specific AUs to the one of the five categories. As all of the previously introduced six datasets include AUs, the same objective labels can be obtained for all of the data. We explore whether additional data and the mappings can be helpful for performance. For further details on the mappings and the protocol, see MEGC2018 \cite{megc2018}.

\begin{table}[htp]
    \centering
    \caption{Task A: holdout-database evaluation, UAR}
    \begin{tabular}{|c|c|c|c|c|}
        \hline
        Method & Extra data & C2 $\rightarrow$ SA & SA $\rightarrow$ C2 & Avg. \\
        \hline
        LBP-TOP \cite{original_lbp_top} $^1$ & & 32.7 & 31.6 & 32.2 \\
        \hline
        3DHOD \cite{3dhog} $^1$ & & 26.9 & 18.7 & 22.8 \\
        \hline
        HOOF \cite{mdmo} $^1$ & & 34.9 & 34.6 & 34.8 \\
        \hline
        ELRCN \cite{elrcn} $^1$ & & 38.2 & 32.2 & 35.2 \\
        \hline
        M2MNet \cite{m2mnet} $^1$ & & 44.0 & 33.7 & 38.9 \\
        \hline
        MER-auGCN \cite{mer_aucgn} $^2$ & & 58.8 & 59.5 & 59.2 \\
        \hline
        RRRN \cite{rrrn} $^3$ & & 55.9 & 59.2 & 57.6 \\
        \hline
        SSSNet \cite{noisy_mer} & & 68.9 & 56.2 & 62.6 \\
        \hline
        ResNet18 \cite{resnet} & & 43.3 & 37.7 & 40.5 \\
        \hline
        Resnet34 \cite{resnet} & & 44.0 & 47.9 & 46.0 \\
        \hline
        Resnet50 \cite{resnet} & & 38.3 & 36.5 & 37.4 \\
        \hline
        RNet18(2+1)D \cite{resnet3d_2p1} & & 20.0 & 22.6 & 21.3\\
        \hline
        \hline
        SSSNet \cite{noisy_mer} & \checkmark & \textbf{83.1} & \textbf{64.0} & \textbf{73.4} \\
        \hline
        ResNet18 \cite{resnet} & \checkmark & 61.1 & 63.6 & 62.3 \\
        \hline
        Resnet34 \cite{resnet} & \checkmark & 67.9 & 56.6 & 62.3 \\
        \hline
        Resnet50 \cite{resnet} & \checkmark & 63.0 & 63.6 & 63.3 \\
        \hline
        RNet18(2+1)D \cite{resnet3d_2p1} & \checkmark & 32.9 & 44.9 & 41.4 \\
        \hline
    \end{tabular}
    \label{tab:megc2018a}
    \\
    \vspace{0.1cm}
    \footnotesize{Results marked with $^1$ were taken from \cite{megc2018}, $^2$ were taken from \cite{mer_aucgn} and $^3$ from \cite{rrrn}.}
\end{table}

\begin{table}[htp]
    \centering
    \caption{Task B: composite dataset evaluation, $F1_{macro}$}
    \begin{tabular}{|c|c|c|}
        \hline
        Method & Extra data & $F1_{macro}$ \\
        \hline
        LBP-TOP \cite{original_lbp_top} $^1$ & & 40.0 \\
        \hline
        3DHOG \cite{3dhog} $^1$ & & 27.1 \\
        \hline
        HOOF \cite{mdmo} $^1$ & & 40.4 \\
        \hline
        ELRCN \cite{elrcn} $^1$ & & 39.3 \\
        \hline
        M2MNet \cite{m2mnet} $^1$ & & 63.9 \\
        \hline
        MER-auGCN \cite{mer_aucgn} $^2$ & & 68.5 \\
        \hline
        RRRN \cite{rrrn} $^3$ & & 66.3 \\
        \hline
        SSSNet \cite{noisy_mer} & & 69.8 \\
        \hline
        ResNet18 \cite{resnet} & & 49.9 \\
        \hline
        Resnet34 \cite{resnet} & & 57.6 \\
        \hline
        Resnet50 \cite{resnet} & & 43.4 \\
        \hline
        RNet18(2+1)D \cite{resnet3d_2p1} & & 24.7 \\
        \hline
        \hline
        SSSNet \cite{noisy_mer} & \checkmark & 69.3 (-0.5) \\
        \hline
        ResNet18 \cite{resnet} & \checkmark & 63.1 (+13.2) \\
        \hline
        Resnet34 \cite{resnet} & \checkmark & 68.1 (+10.5) \\
        \hline
        Resnet50 \cite{resnet} & \checkmark & 64.6 (+21.2) \\
        \hline
        RNet18(2+1)D \cite{resnet3d_2p1} & \checkmark & 48.2 (+23.5) \\
        \hline
    \end{tabular}
    \label{tab:megc2018b}
\end{table}

\textbf{Results} From Table \ref{tab:megc2018a} we can see the results for the holdout-database evaluation using UAR (unweighted average recall) \cite{megc2018}. The third column shows the result when training on CASME II and then testing on SAMM, and vice versa for the fourth column. Extra data refers to using the remaining five datasets during the training. We can see that without additional data SSSNet \cite{noisy_mer} achieves the best result, while the larger ResNets fall behind in terms of performance. However, with additional data SSSNet achieves the highest performance by a sizeable margin.

The results for the composite dataset evaluation can be seen from Table \ref{tab:megc2018b}. With no additional data the performance between MER-auGCN and SSSNet are highly similar. The use of additional data does not improve the performance for SSSNet, but does so for the larger ResNet models. Compared to the holdout-database evaluation the additional data does not bring benefits to SSSNet's performance. We postulate that the reason for this is the simpler nature of the composite dataset task and the access to similar training data due to the use of LOSO. As larger models with more parameters require more data, the limited data did not suffice and hence the additional data can still bring an increase for the ResNet models.



\ifCLASSOPTIONcompsoc
  \section*{Acknowledgments}
\else
  \section*{Acknowledgment}
\fi

This work was supported by the Academy of Finland for Academy Professor project EmotionAI (grants 336116, 345122), the University of Oulu \& The Academy of Finland Profi 7 (grant 352788), and Ministry of Education and Culture of Finland for AI forum project.

\ifCLASSOPTIONcaptionsoff
  \newpage
\fi



%
\AtNextBibliography{\footnotesize}
\printbibliography
%
%

%

\begin{IEEEbiography}[{\includegraphics[width=1in,height=1.25in,clip,keepaspectratio]{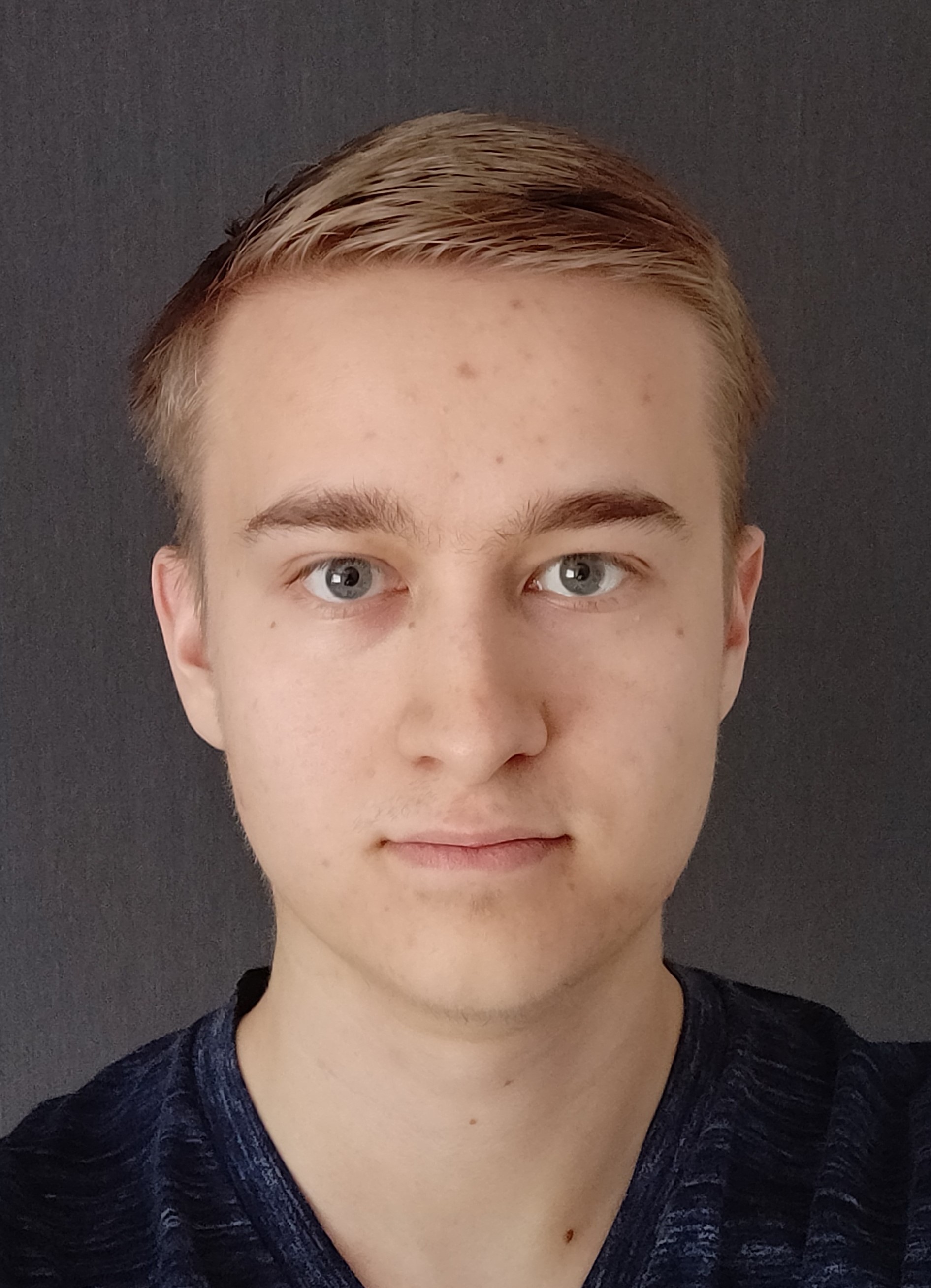}}]{Tuomas Varanka}
is currently a doctoral researcher with the Center for
Machine Vision and Signal Analysis, University of Oulu, Oulu, Finland.
He received his B.S. and M.S. degree in computer science and engineering from the University of Oulu in 2019 and 2020, respectively. His current research
interests include affective computing, facial and micro-expression analysis. 
\end{IEEEbiography}

\begin{IEEEbiography}[{\includegraphics[width=1in,height=1.25in,clip,keepaspectratio]{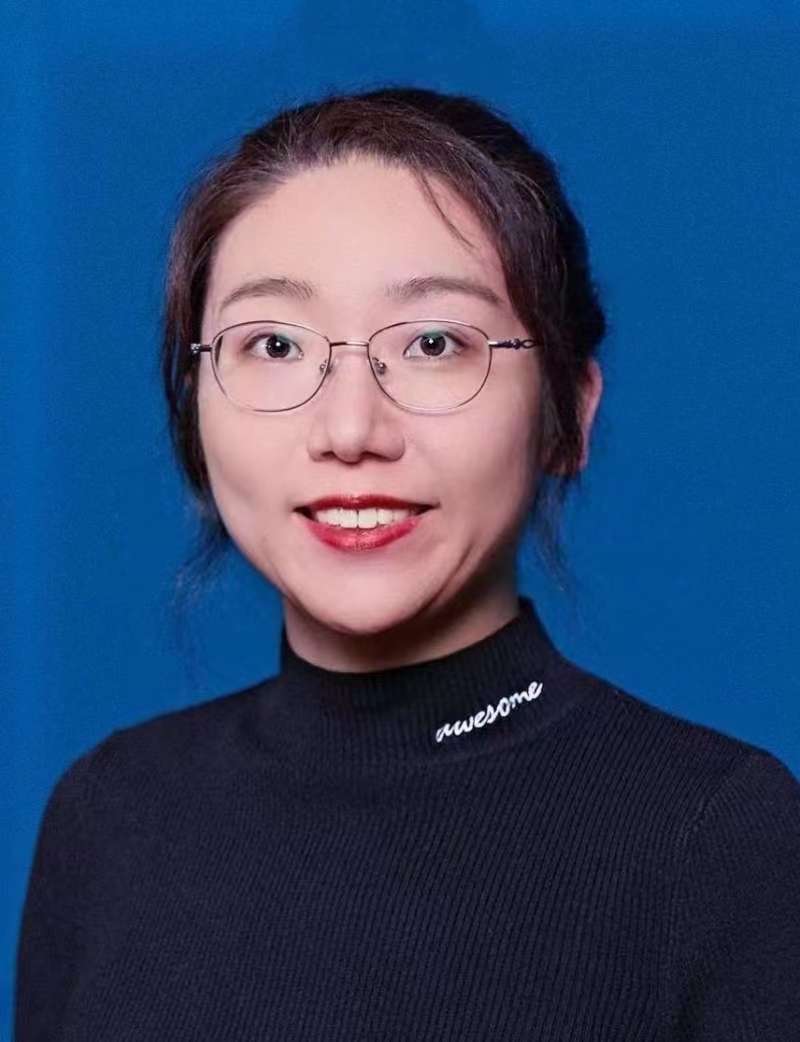}}]{Yante Li}
received her Ph.D. degree in Computer Science from University of Oulu, Oulu, Finland, in 2022. She is currently a Postdoc in the Center for Machine Vision and Signal Analysis of University of Oulu. Her current research interests include affective computing, micro-expression analysis and facial action unit detection.
\end{IEEEbiography}
\begin{IEEEbiography}[{\includegraphics[width=1in,height=1.25in,clip,keepaspectratio]{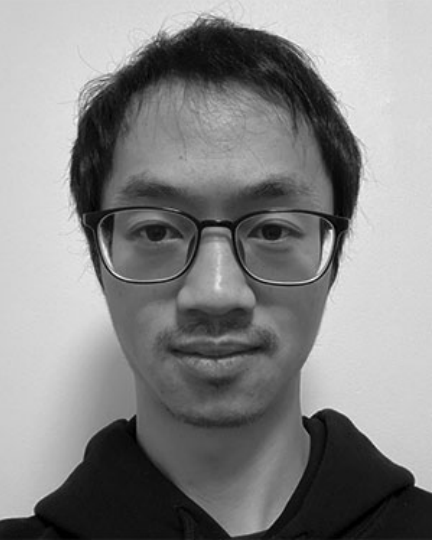}}]{Wei Peng}
is currently Postdoctoral Scholar at Stanford University. He received his Ph.D. degree from University of Oulu, Finland, in 2022. He received the M.S. degree in computer science from the Xiamen University, Xiamen, China, in 2016. His articles have published in mainstream conferences and journals, such as CVPR, AAAI, ICCV, ACM Multimedia, TPAMI, and TIP. His current research interests include machine learning, medical imaging analysis, and neuroscience.
\end{IEEEbiography}
\begin{IEEEbiography}[{\includegraphics[width=1in,height=1.25in,clip,keepaspectratio]{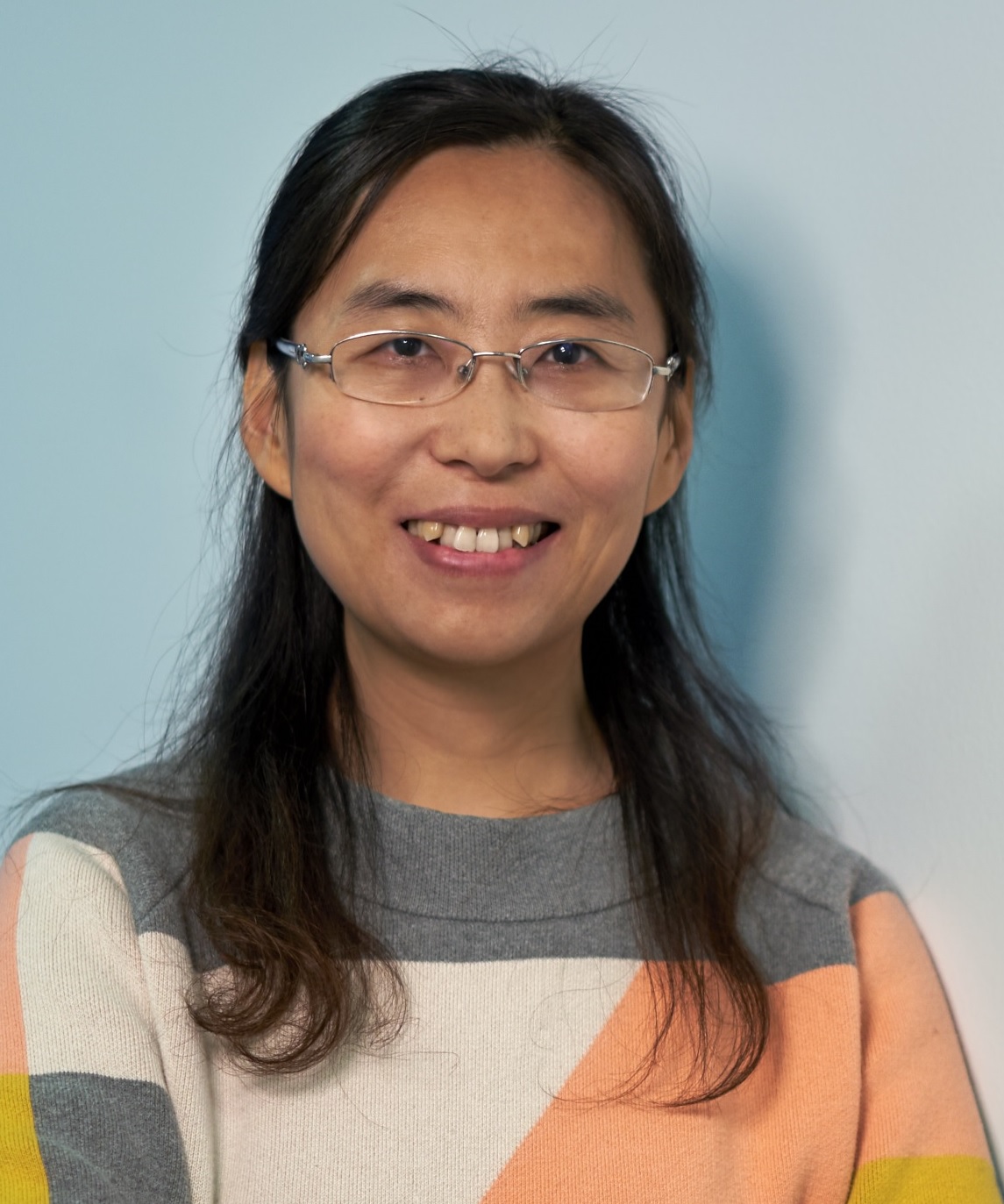}}]{Guoying Zhao}
(IEEE Fellow 2022) received the Ph.D. degree in computer science from the Chinese Academy of Sciences, Beijing, China, in 2005. She is currently an Academy Professor and full Professor (tenured in 2017) with University of Oulu. She is also a visiting professor with Aalto University. She is a member of Finnish Academy of Sciences and Letters, IAPR Fellow and AAIA Fellow. She has authored or co-authored more than 300 papers in journals and conferences with 21000+ citations in Google Scholar and h-index 68. She is panel chair for FG 2023, publicity chair of 22nd Scandinavian Conference on Image Analysis (SCIA 2023), was co-program chair for ACM International Conference on Multimodal Interaction (ICMI 2021), co-publicity chair for FG2018, and has served as area chairs for several conferences and was/is associate editor for IEEE Trans. on Multimedia, Pattern Recognition, IEEE Trans. on Circuits and Systems for Video Technology, and Image and Vision Computing Journals. Her current research interests include image and video descriptors, facial-expression and micro-expression recognition, emotional gesture analysis, affective computing, and biometrics. Her research has been reported by Finnish TV programs, newspapers and MIT Technology Review.
\end{IEEEbiography}






\end{document}